\documentclass[conference]{IEEEtran}

\usepackage{cite}
\usepackage{amsmath,amssymb,amsfonts}
\usepackage{algorithmic}
\usepackage{graphicx}
\usepackage{textcomp}
\usepackage{xcolor}
\usepackage{booktabs}
\usepackage{balance}
\usepackage{threeparttable}
\usepackage{tabularx}
\newcolumntype{Y}{>{\centering\arraybackslash\hsize=1.8\hsize}X}
\newcolumntype{Z}{>{\centering\arraybackslash\hsize=0.7\hsize}X}
\def\BibTeX{{\rm B\kern-.05em{\sc i\kern-.025em b}\kern-.08em
    T\kern-.1667em\lower.7ex\hbox{E}\kern-.125emX}}
\begin{document}

\title{When More Is Not Better: Component Anti-Synergy in a P300 Speller}

\author{\IEEEauthorblockN{Lucas Yang}
\IEEEauthorblockA{\textit{Parkland High School}\\
Allentown, PA, USA \\
yanglucas2028@gmail.com}
\and
\IEEEauthorblockN{Rui Liu}
\IEEEauthorblockA{\textit{Dept. of Computer Science} \\
\textit{Stony Brook University}\\
Stony Brook, NY, USA \\
ruiliu1@cs.stonybrook.edu}
\and
\IEEEauthorblockN{Fusheng Wang}
\IEEEauthorblockA{\textit{Dept. of Computer Science} \\ \textit{Dept. of Biomedical Informatics} \\
\textit{Stony Brook University}\\
Stony Brook, NY, USA \\
fusheng.wang@stonybrook.edu}
}

\maketitle

\begin{abstract}
P300 brain-computer interface (BCI) spellers can provide hands-free communication for people with severe motor impairments. Modern pipelines combine multiple individually promising components, often assuming that “more-is-better.” We tested this assumption using a four-component full-factorial experiment varying the inclusion of Euclidean Alignment (EA), xDAWN spatial filtering, subject calibration, and language model priors on a public P300 dataset. Performance was evaluated using accuracy, repetitions, and information transfer rate (ITR) with mixed-effects models. Results show that the value of components is conditional rather than additive. Calibration was the strongest singular contributor, while EA compensated for its absence in zero-calibration settings. Adding independently useful components could also reduce performance, revealing \textit{component anti-synergy}. Contrary to conventional wisdom, LM support was not universally beneficial: its effect depends strongly on the strength of the underlying EEG pipeline, while results from a larger LM showed a similar pattern. Together, these findings challenge maximal ``all-on" pipeline design and highlight the value of selecting spatial and language-support components according to the quality of available EEG evidence. 
\end{abstract}


\section{Introduction}
For people with severe communication and motor impairments, including individuals with amyotrophic lateral sclerosis (ALS), loss of speech and movement can make conventional input devices difficult or impossible to use \cite{nijboer2008p300}. Brain–computer interface (BCI) spellers offer a hands-free alternative by translating brain activity into character selections \cite{sellers2006p300}. In a P300 speller using the Farewell-Donchin grid, rows or columns of a character matrix flash while the user attends to a target. The resulting P300 event-related potential (ERP) in the electroencephalography (EEG) signal is used to infer the intended character \cite{farwell1988talking}. Practical systems must balance communication accuracy and efficiency, system calibration complexities, and robustness across distinct users.

Modern P300 pipelines combine multiple components to address these goals: Euclidean Alignment (EA) supports cross-subject/session transfer, xDAWN (XD) enhances P300 spatial representation, subject calibration (CB) personalizes the decoder, and language models (LM) offer contextual priors\cite{he2020transfer, rivet2009xdawn, kindermans2012masses, speier2014integrating}. Although each one has demonstrated value, prior studies typically evaluate individual modules or aggregate pipeline performance, leaving their interactions unclear. Individually useful components may  become complementary, redundant, or interfering when combined.

This uncertainty is especially notable for zero- or low-calibration settings, which reduce setup burden but also limit subject-specific adaptability \cite{kindermans2014integrating} and require other components to compensate for missing calibration. Moreover, although LM and, more recently, large language model (LLM) priors may improve contextual prediction, it remains unclear whether their effects depend on the strength of the underlying EEG evidence shaped by other pipeline components \cite{speier2014integrating, hong2026chatbci}. We therefore ask: (1) Are component effects additive, redundant, or anti-synergistic? (2) How do component behaviors differ with versus without calibration, and can any partially compensate for its absence? (3) How do LM and LLM priors behave when EEG decoding is well-aligned, personalized, or weak?

We address these questions using a public P300 matrix-speller dataset containing 10 subjects across three sessions \cite{arico2014influence, jayaram2018moabb}. During processing, EA, xDAWN, calibration, and LM support were independently enabled or disabled in a $2^4=16$ full-factorial design. Performance was evaluated using accuracy, repetitions per character, and information transfer rate (ITR) with linear mixed-effects models. Robustness was assessed using a larger LM, an alternative random-effects specification, and cross-session curation.

Our contributions are:
\begin{itemize}
    \setlength{\itemsep}{0pt}
    \setlength{\parskip}{0pt}
    \setlength{\parsep}{0pt}
    \item We show through full-factorial analysis that P300 component value is conditional rather than additive and identify configurations with redundancy penalties.
    \item We clarify zero-calibration design: calibration is the dominant contributor, while xDAWN benefits uncalibrated decoding only when paired with EA and can reduce efficiency once EA and calibration are both present.
    \item We establish a boundary condition for LM/LLM support: language priors are harmful when EEG evidence is insufficiently aligned or personalized, whereas calibration largely neutralizes their interference.
\end{itemize}

\begin{figure*}[hbt]
    \centering
    \includegraphics[width=0.98\textwidth]{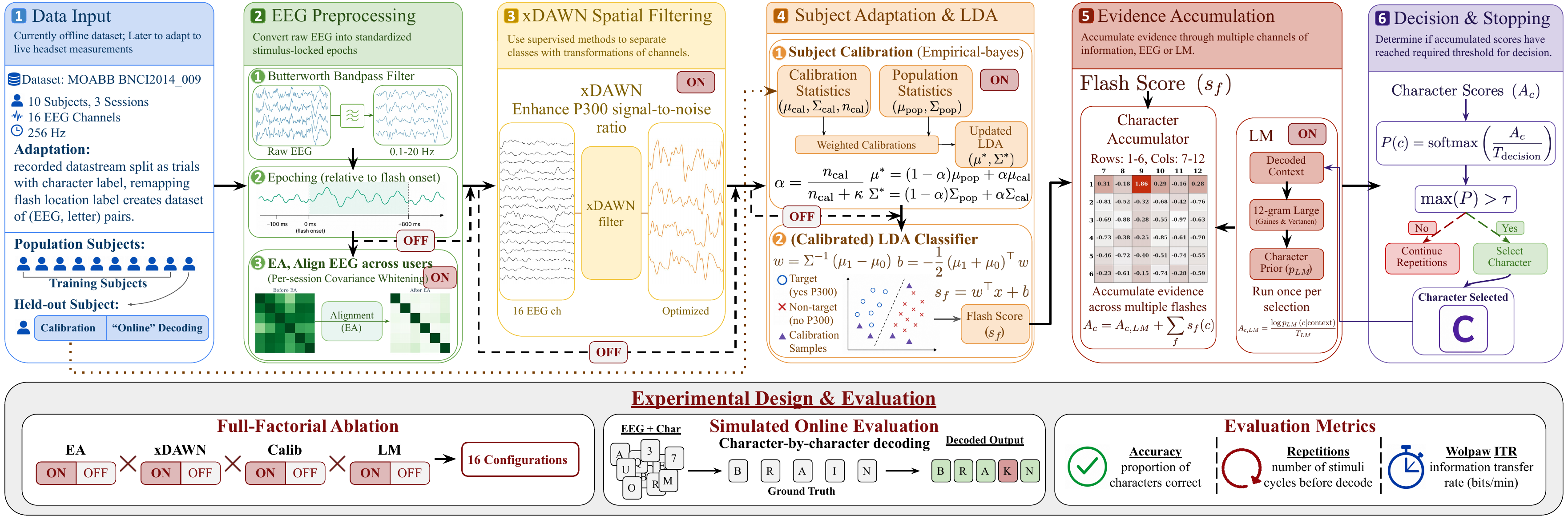}
    \caption{P300 Speller Pipeline and Full-Factorial Design}
    \label{fig:framework}
\end{figure*}
\section{Related Work}
Previous P300 research has established benefits for components in different stages of the decoding pipeline. However, individual-component studies estimate stand-alone benefits, whereas complete-system studies emphasize aggregate performance. Neither directly reveals whether adding one component changes the marginal value of another.

These interactions are particularly important for zero-calibration decoding, where successful systems may rely on compensatory mechanisms elsewhere in the pipeline. For example, Kindermans et al. combined inter-subject transfer, continuous adaptation, language modeling, and dynamic stopping \cite{kindermans2014integrating}. EA and xDAWN may exhibit a similar compensatory interaction: EA improves cross-subject ERP transfer without labeled target-user data \cite{he2020transfer}, while xDAWN enhances P300 spatial separation \cite{li2020transfer}. Yet it remains unclear whether EA can compensate for missing calibration, whether it is needed for xDAWN to remain beneficial under zero calibration, or whether the two become redundant once calibration is available.

A similar ambiguity applies to language support. Studies reporting benefits from LM- or LLM-assisted P300 spelling may have generally used trained or calibrated EEG decoders before evaluating language assistance: Oken et al. calibrated participants before LM-EEG fusion, while Speier et al. used training trials before online LM evaluation \cite{speier2018improving, oken2014braincomputer}. Consequently, the independent effect of an LM under unaligned or unpersonalized EEG remains unclear, and a detrimental effect could be masked by stronger upstream decoding. These observations motivate jointly testing EA, xDAWN, calibration, and language support on their individual and interaction effects.

\section{Methods}
\subsection{Dataset and Simulated Online Decoding}
We evaluated the P300 speller using the public BNCI2014\_009 dataset distributed through the Mother of All BCI Benchmarks (MOABB) framework \cite{arico2014influence,jayaram2018moabb}. The dataset contains EEG from 10 subjects across three sessions from 16 channels at 256 Hz during a visual P300 matrix-speller task. Recorded EEG was reorganized for simulated online evaluation. During decoding, characters were processed sequentially, allowing previously decoded text to provide LM context while preserving the original EEG responses associated with each stimulus presentation. The primary analyses evaluated decoding within session, with cross-session decoding examined as a robustness test. 

\subsection{P300 Speller Pipeline}
The decoding pipeline is summarized in Fig. \ref{fig:framework}. Raw EEG was bandpass filtered (0.1-20 Hz) and epoched from  $-100$ to $+800$ ms around each flash. Depending on the experimental condition, Euclidean Alignment (EA) \cite{he2020transfer} and xDAWN spatial filtering \cite{rivet2009xdawn} were optionally applied before linear discriminant analysis (LDA) classification.

With calibration, the population LDA was adapted using empirical-Bayes shrinkage that blended population and subject-specific statistics \cite{kindermans2012masses} with weight,
\[
\alpha=\frac{n_{\mathrm{cal}}}{n_{\mathrm{cal}}+\kappa}, \tag{1}
\]
where more calibration data increased subject-specific weighting \cite{kindermans2014integrating}. Without calibration, the population model was used directly.

LDA evidence accumulated across repeated flashes for each candidate character. When enabled, the LM generated a character-level prior from the previously decoded text context \cite{speier2014integrating}. This prior was introduced once at the beginning of each character selection and combined with subsequently accumulated EEG evidence \cite{speier2018improving}. After each repetition, character scores were converted to softmax probabilities; decoding stopped when the maximum probability exceeded the selection threshold; otherwise, another repetition followed \cite{kindermans2014integrating}. 

\subsection{Full-Factorial Design and Evaluation Measures}
EA, xDAWN (XD), calibration (CB), and LM were independently enabled or disabled, yielding $2^4=16$ configurations while all other procedures remained fixed.

Performance was assessed using three outcomes: \textit{accuracy}, the proportion of correctly decoded characters; \textit{repetitions}, the mean number of stimulus repetitions per character, with lower value indicating greater efficiency; and \textit{information transfer rate (ITR)}, measured in bits/min and jointly reflecting accuracy and efficiency using Wolpaw's definition \cite{wolpaw1998eeg}:
\[
B=\log_2N+A\log_2A+(1-A)\log_2\left(\frac{1-A}{N-1}\right) \tag{2}
\]\[
\mathrm{ITR}=\frac{60B}{Rt} \tag{3}
\]
where $B$ is bits/selection, $A$ is accuracy, $N$ the number of choices, $R$ repetitions per selection, and $t$ seconds per repetition.

\subsection{Primary Mixed-Effects Analysis}
Within-session outcomes were analyzed separately using linear mixed-effects models with a subject random intercept. For outcome $Y_{ijt}$ from subject $i$, sentence $j$, and configuration $t$, the four factors (CB, EA, XD, and LM) $X_{kt}$ were effect-coded as $-1$ (off) and $+1$ (on). The model is

\[
\begin{aligned}
Y_{ijt}={}&\beta_0
+\sum_{k=1}^{4}\beta_kX_{kt}
+\sum_{k<\ell}\beta_{k\ell}X_{kt}X_{\ell t}\\
&+\sum_{k<\ell<m}\beta_{k\ell m}X_{kt}X_{\ell t}X_{mt} \\
&+\beta_{1234}\prod_{k=1}^{4}X_{kt}
+u_i+\epsilon_{ijt},
\end{aligned}
\tag{4}
\]
where the $\beta$ terms represent factorial effects, $u_i\sim N(0,\sigma_u^2)$ is the subject random intercept, and $\epsilon_{ijt}\sim N(0,\sigma^2)$ is residual error. The four-way term preserved model hierarchy, while interpretation emphasized lower-order interactions. FDR-adjusted $p<.05$ indicated significance.

\subsection{Robustness Check}
Three robustness tests were conducted. First, we re-estimated the within-session models with subject-and-sentence random intercepts, adding a sentence random intercept to account for variation across text items. Second, we applied the primary subject-random-intercept model to cross-session accuracy, repetitions, and ITR to test generalization under session transfer. Third, we replaced the primary LM with a larger contemporary LLM \cite{hong2026chatbci} and repeated the within-session analysis using the primary mixed-effects specification. Together, these tests evaluated robustness to the random-effects structure, session transfer, and language-prior choice.

\begin{figure*}[hbt]
    \centering
    \includegraphics[width=0.98\textwidth]{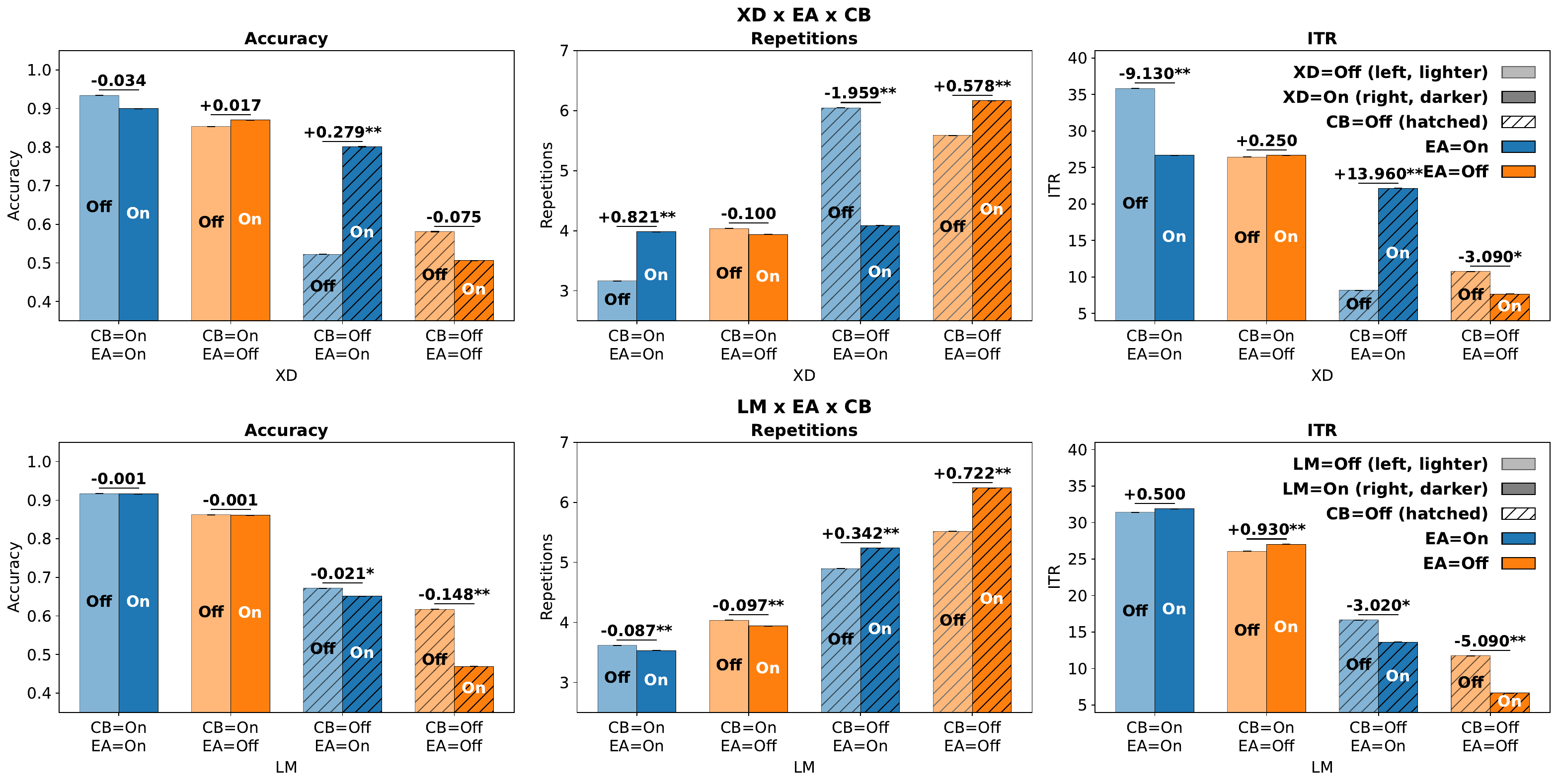}
    \caption{xDAWN and LM Effects Conditional on Calibration and Euclidean Alignment}
    \vspace{3pt}
    \scriptsize{Note: $^{*}p_{\mathrm{FDR}}<.05$; $^{**}p_{\mathrm{FDR}}<.005$}
    \label{fig:six_plot}
\end{figure*}

\section{Results}
\subsection{Main Effects}
Table \ref{tab:mixedlm_estimates} summarizes the estimated effects and FDR-adjusted $p$-values from the primary within-session mixed-effects models. Calibration produced the largest favorable main effects across accuracy, repetitions, and ITR, followed by EA. xDAWN showed no significant average main effect, whereas LM support significantly worsened all three outcomes. These contrasting main effects motivated examination of component interactions and possible anti-synergy. 

\begin{table}[thb]
\centering
\footnotesize
\caption{Mixed-Effects Model Estimates for Word-Level Outcomes}
\label{tab:mixedlm_estimates}
\begin{tabular}{lccc}
\toprule
Contrast & Accuracy & Repetition & ITR \\
\midrule
\multicolumn{4}{l}{\textit{Main Effects}} \\
CB                & 0.288**  & -20.158*  & 16.728** \\
EA                & 0.085**  & -7.621*   & 5.654**  \\
XD                & 0.045    & -1.685    & 0.601    \\
LM                & -0.035*  & 3.455**   & -2.013*  \\
\midrule
\multicolumn{4}{l}{\textit{Two-way Interactions}} \\
XD $\times$ CB    & -0.115** & 12.908**  & -9.624** \\
XD $\times$ EA    & 0.155*   & -9.903*   & 4.361*   \\
LM $\times$ CB    & 0.086**  & -7.191**  & 4.369**  \\
LM $\times$ EA    & 0.061*   & -2.780*   & 1.089    \\
XD $\times$ LM    & 0.064*   & -1.695    & 1.037    \\
CB $\times$ EA    & -0.058   & 4.655     & -0.428   \\
\midrule
\multicolumn{4}{l}{\textit{Three-way Interactions}} \\
XD $\times$ CB $\times$ EA & -0.410* & 41.800** & -26.148** \\
LM $\times$ CB $\times$ EA & -0.118* & 4.411*   & -1.650    \\
XD $\times$ LM $\times$ CB & -0.148* & 3.948    & -2.466    \\
XD $\times$ LM $\times$ EA & -0.164** & 15.314** & -8.872** \\
\midrule
\multicolumn{4}{l}{\textit{Four-way Interaction}} \\
XD $\times$ LM $\times$ CB $\times$ EA & 0.320** & -30.274** & 15.836** \\
\bottomrule
\end{tabular}

\vspace{5pt}
\raggedright
\scriptsize
Note: Subject-random-intercept mixed-effects models on with-session outcomes. CB = calibration, EA = Euclidean Alignment, XD = xDAWN, and LM = language model. Negative estimates indicate fewer repetitions but lower accuracy/ITR. FDR-corrected significance: * $p<0.05$, ** $p<0.01$, *** $p<0.001$.
\end{table}

\subsection{Interaction Effects}
Table \ref{tab:mixedlm_estimates} shows that component effects were conditional: xDAWN interacted with both calibration and EA across all outcomes, while LM interacted strongly with calibration. The focal three-way interactions are shown in Fig.~\ref{fig:six_plot}, illustrating the effects of enabling xDAWN and LM on accuracy, repetitions, and ITR across EA and calibration conditions. Values represent component-on minus component-off mean differences.

\subsubsection{xDAWN Anti-Synergy Depends on Calibration and EA}
The EA $\times$ xDAWN $\times$ calibration interaction was significant across all three outcomes. Without calibration, xDAWN improved accuracy/ITR by $+0.279/+13.96$ when paired with EA, but worsened them by $-0.075/-3.09$ without EA; repetitions changed by $-1.959$ versus $+0.578$. After calibration, xDAWN added little without EA ($+0.017$ accuracy, $+0.25$ ITR) and reduced efficiency significantly when EA was also enabled ($+0.821$ repetitions, $-9.13$ ITR). Thus, xDAWN was beneficial in an EA-aligned zero-calibration pipeline but showed a clear redundancy penalty after alignment and calibration were both present. 

\subsubsection{LM Interference Depends on EEG Support}
The EA $\times$ LM $\times$ calibration interaction was significant for accuracy and repetitions, but not ITR. Without calibration, LM harm was greatest without EA (accuracy/ITR: $-0.148/-5.09$) and smaller with EA ($-0.021/-3.02$); repetitions increased by $0.722$ and $0.342$. After calibration, LM effects were nearly neutral across EA conditions (accuracy $-0.001/-0.001$; repetitions $-0.097/-0.087$; ITR $+0.93/+0.50$). Thus, EA partially mitigated LM interference under zero calibration, while calibration largely neutralized it.

\subsection{Robustness Analyses}
The robustness tests largely preserved the core findings (Table~\ref{tab:robustness_tests}). First, adding a sentence random intercept retained the focal xDAWN interaction across all outcomes and the LM interaction for accuracy and repetitions. Second, cross-session analysis again supported xDAWN's dependence on EA and calibration. It also retained the negative LM main effect and the LM $\times$ calibration interaction across all outcomes, although the EA $\times$ LM $\times$ calibration interaction remained significant only for repetitions. Third, replacing the LM with a larger LLM reproduced the negative language-prior main effect and its attenuation by calibration; its three-way interaction was significant for accuracy. Overall, the most stable results were calibration's dominant benefit, xDAWN's dependence on EA under zero calibration, the negative average effect of language priors, and calibration's mitigation of that interference.

\begin{table*}[thb]
\centering
\footnotesize
\caption{Robustness Tests for Mixed-Effects Model Estimates}
\label{tab:robustness_tests}
\resizebox{\textwidth}{!}{%
\begin{tabular}{lccccccccc}
\toprule
& \multicolumn{3}{c}{Robustness Test 1: Alternative Mixed-Effects Model} & \multicolumn{3}{c}{Robustness Test 2: Cross-Session Outcomes} & \multicolumn{3}{c}{Robustness Test 3: Alternative LLM Model} \\
\cmidrule(lr){2-4} \cmidrule(lr){5-7} \cmidrule(lr){8-10}
Contrast & Accuracy & Repetition & ITR & Accuracy & Repetition & ITR & Accuracy & Repetition & ITR \\
\midrule
\multicolumn{10}{l}{\textit{Main Effects}} \\
CB & 0.288*** & -20.158*** & 16.728*** & 0.235 & -20.158* & 14.987* & 0.287** & -20.307* & 16.926** \\
EA & 0.085*** & -7.621*** & 5.654*** & 0.091** & -7.477** & 5.169** & 0.087** & -7.345* & 5.518** \\
XD & 0.045*** & -1.685*** & 0.601* & 0.044* & -3.256** & 1.123 & 0.046 & -1.980 & 0.696 \\
LM & -0.035*** & 3.455*** & -2.013*** & -0.075** & 3.875** & -3.319** & -0.043** & 2.643** & -1.668** \\
\midrule
\multicolumn{10}{l}{\textit{Two-way Interactions}} \\
XD $\times$ CB & -0.115*** & 12.908*** & -9.624*** & -0.093** & 9.985** & -7.118** & -0.111** & 12.608** & -9.470** \\
XD $\times$ EA & 0.155*** & -9.903*** & 4.361*** & 0.167** & -9.373* & 4.401 & 0.152* & -9.700* & 4.232* \\
LM $\times$ CB & 0.086*** & -7.191*** & 4.369*** & 0.087** & -7.905** & 3.909* & 0.084** & -7.489** & 4.765** \\
LM $\times$ EA & 0.061*** & -2.780** & 1.089 & 0.064* & -2.804* & 1.288 & 0.064* & -2.226 & 0.818 \\
\midrule
\multicolumn{10}{l}{\textit{Three-way Interactions}} \\
XD $\times$ CB $\times$ EA & -0.410*** & 41.800*** & -26.148*** & -0.408** & 41.428* & -24.733** & -0.404 & 41.488** & -25.618** \\
LM $\times$ CB $\times$ EA & -0.118*** & 4.411** & -1.650 & -0.129 & 5.223* & -1.424 & -0.127* & 4.675 & -2.503 \\
\bottomrule
\end{tabular}%
}
\raggedright

\vspace{2pt}
\scriptsize{
Note: Only focal interactions are reported. Tests 1 and 3 use within-session outcomes; Tests 2 and 3 use a subject-random-intercept model. Tests 1--2 use the LM, and Test 3 uses the LLM. Significance (FDR-corrected): * $p<0.05$, ** $p<0.01$, *** $p<0.001$.}
\end{table*}

\section{Discussion}
Our key contribution is the identification of \textit{component anti-synergy}: components that are beneficial individually can become redundant or even harmful when combined. Calibration was the dominant contributor, while EA enabled xDAWN to improve zero-calibration decoding. Once both EA and calibration were present, however, xDAWN increased repetitions and reduced ITR. A plausible explanation is that EA and calibration already improve the signal representation through alignment and personalization, leaving xDAWN with little additional value while adding redundant spatial filtering \cite{he2020transfer,rivet2009xdawn}. Future work should test this mechanism directly.

\begin{table}[htb]
\centering
\begin{threeparttable}
\caption{Interaction-Guided Configuration Strategies}
\label{tab:recommendations}
\scriptsize
\setlength{\tabcolsep}{2.5pt}
\renewcommand{\arraystretch}{1.25}
\begin{tabularx}{0.45\textwidth}{@{}l Y Z Z Z@{}}
\hline
Strategy & EA, XD, CB, LM & Acc. & Rep. & ITR \\
\hline
Naive minimalist & (All not included) & 0.731 & 4.82 & 16.1 \\
Naive maximalist & (All included) & 0.905 & 4.02 & 27.3 \\
Calibrated recommendation & (EA, CB) & 0.933 & 3.20 & 35.5 \\
Zero-calib. recommendation & (EA, XD) & 0.819 & 3.67 & 25.3 \\
\hline
\end{tabularx}
\begin{tablenotes}
\scriptsize
\item Configuration order: EA/xDAWN/calibration/LM. Higher accuracy and ITR and lower repetitions are preferable.
\end{tablenotes}
\end{threeparttable}
\end{table}

The language-model results reveal a second important boundary condition. Previous studies reporting benefits from language information generally evaluated trained or calibrated EEG decoders [14], [17]. In contrast, both LM and LLM priors in our study were harmful on average, especially when neither EA nor calibration was present. EA partly mitigated this interference, while calibration largely neutralized it. One possible explanation is that Weak EEG evidence may allow contextual priors to dominate and propagate decoding errors. The larger LLM showed the same pattern, suggesting that model scale alone does not resolve the fusion problem \cite{speier2018improving,hong2026chatbci}. Future systems may therefore benefit from adjusting LM/LLM influence based on EEG quality.

Table~\ref{tab:recommendations} translates the interaction findings into practical configuration guidance. The calibrated recommendation (EA+CB) achieved the best overall performance, with the highest accuracy (0.933) and ITR (35.5 bits/min) and the fewest repetitions (3.20), outperforming both naive baselines. Under zero calibration, EA+xDAWN substantially improved over the minimalist baseline (0.819 vs. 0.731 accuracy; 25.3 vs. 16.1 ITR; 3.67 vs. 4.82 repetitions) and also required fewer repetitions than the maximalist baseline. This suggests a practical low-burden configuration when subject-specific calibration is unavailable. After calibration, retaining EA while removing xDAWN avoids the observed efficiency penalty. Overall, these results support selective rather than maximal component configuration. 

Limitations include the use of a single public dataset with simulated online EEG recording. The system was also not tested with real ALS patients, a key target population for P300 communication systems. Future work should therefore validate the observed interactions across additional datasets and headsets, and test genuine online use with clinical participants.

\section{Conclusion}
We evaluated EA, xDAWN, subject calibration, and LM support across 16 full-factorial configurations. Three key findings emerged: component anti-synergy, where adding individually useful modules can reduce performance; negative language-model effects, particularly when neither calibration nor EA is present; and EA as a potential compensatory mechanism under zero calibration, especially by enabling xDAWN. Together, these results suggest potential value in conditionally configuring P300 pipelines, with spatial and language support adapted to the available EEG evidence.

\section*{Acknowledgment}
The authors would like to thank Dr. Fusheng Wang and the CSIRE program at Stony Brook University for supporting this research.

\balance
\bibliographystyle{IEEEtran}
\bibliography{reference}

\end{document}